%% file: neurips_2026.tex
\pdfoutput=1
\documentclass{article}

\PassOptionsToPackage{numbers}{natbib}

 \usepackage[main, final]{neurips_2026}

\usepackage[utf8]{inputenc} %
\usepackage[T1]{fontenc} %
\usepackage{hyperref} %
\usepackage{url} %
\usepackage{booktabs} %
\usepackage{amsfonts} %
\usepackage{nicefrac} %
\usepackage{microtype} %
\usepackage{xcolor} %

\usepackage{enumitem}
\usepackage{amsmath} 
\usepackage{graphicx}
\usepackage{array}
\usepackage{xcolor}
\usepackage{multirow}
\usepackage[table]{xcolor}
\usepackage{arydshln}
\usepackage{placeins}

\title{\textsc{XBridge}: Entity-Grounded Latent Bridge for Heterogeneous LLM Communication}

\author{%
  \textbf{Wooseong Yang$^{1}$ \enspace Wei-Chieh Huang$^{1}$ \enspace Weizhi Zhang$^{1}$ \enspace Yu Wang$^{2}$} \\[3pt]
  \textbf {Philip S.~Yu$^{1, \dagger}$ \enspace Junhyun Lee$^{3,4,\dagger}$} \\[3pt]
  {\small $^{1}$University of Illinois Chicago \ \
  $^{2}$Capital One AI Foundations \ \ }\\
  {\small $^{3}$Hankuk University of Foreign Studies \ \
  $^{4}$Noah's Farm \quad }\\[1pt]
  {\small $^{\dagger}$Corresponding authors}
}

\begin{document}

\maketitle

\input{sections/0_abstract}
\input{sections/1_introduction}
\input{sections/2_motivation}

\input{sections/3_method}
\input{sections/4_experiments}

\input{sections/7_conclusion}

\input{sections/limitation}

\bibliography{references}
\bibliographystyle{unsrtnat}
\input{sections/x_appendix}
\end{document}

%% file: sections/0_abstract.tex
\begin{abstract}
Heterogeneous multi-agent LLM systems, where agents are powered by different model families, can outperform homogeneous configurations by reducing redundant reasoning patterns. 
Yet existing communication protocols either operate through text, discarding the sender's internal representations, or require architectural homogeneity for latent-level transfer. We identify the \textit{entity grounding problem} in cross-architecture communication: cross-attention bridges that transfer continuous representations across different LLM families suffer from rare-token compression collapse, where entity identity is lost in the continuous bottleneck (bridge-only F1 $\sim$30\%). We propose \textsc{XBridge}, a decode-free communication protocol that addresses this through two mechanisms. Lexical Anchor Mapping (LAM) maps the sender's original context tokens to the receiver's vocabulary, providing discrete entity anchors. A Latent Enrichment Bridge (LEB) lets the receiver query the sender's hidden states for contextual enrichment. The entity anchors ground the bridge's contextual signals to specific entities through the receiver's own self-attention. 
Across three model families (Llama, Qwen, and Mistral), seven benchmarks, and both communication directions, \textsc{XBridge} outperforms text-based communication on all seven tasks for each model pair while achieving 11$\times$ lower latency, and in a same-architecture setting it also exceeds a KV-sharing baseline on six of seven tasks.
LEB requires only 264M trainable parameters (3.8\% of the receiver), is trained on a small balanced sample set, and adds negligible inference overhead.
Our implementation is available at \url{https://github.com/WooseongYang/XBridge}.
\end{abstract}

%% file: sections/1_introduction.tex
\section{Introduction}
\label{sec:introduction}

Multi-agent systems (MAS) can improve the reasoning capabilities of large language models (LLMs) by distributing computation across multiple inference passes \citep{zou2025latentmas}. 
Recent findings show that when interacting agents share identical parameters and inductive biases, their interactions may degenerate into redundant majority voting rather than surfacing complementary reasoning \citep{hu2026multiagent,choi2026debate}. 
A natural way to mitigate this redundancy is to use heterogeneous ensembles: configurations comprising distinct model families with diverse tokenizers, training corpora, architectures, and alignment procedures. 
Yet this representational diversity introduces a severe communication bottleneck: agents must exchange information across structurally incompatible latent spaces.

Existing communication paradigms face an efficiency-fidelity trade-off. 
Text-based methods, such as natural language communication (NLComm) \citep{du2024debate}, are architecture-agnostic but incur autoregressive decoding latency and compress the sender's internal state into a textual summary. 
Latent-level methods avoid sender-side decoding, but existing approaches such as KV cache sharing \citep{shi2026kvcomm} and hidden-state injection \citep{ramesh2025activation} typically assume architectural compatibility, including aligned hidden dimensions, attention layouts, positional encodings, or tokenizers. 
Learned continuous projections offer a possible route across model families \citep{fu2025cache}.
However, we find that continuous projections alone degrade in a specific and practically important way: they can transfer contextual information without reliably preserving the entities to which that context refers.

\input{figures/fig_hero}

We define this degradation as the \textit{entity grounding problem}. 
In heterogeneous LLM communication, a continuous bridge can transfer contextual information from the sender, but it struggles to preserve the discrete entity identities, such as names, numbers, or rare tokens, to which that context refers.
We use \textit{entity fidelity} to refer to preserving the correct lexical identity, and \textit{entity grounding} to refer to binding the sender's contextual signal to the corresponding receiver-native lexical anchor.
In bridge-only experiments, the receiver recovers aspects of the context but often loses the exact identities of the participating entities.
Homogeneous communication settings are less exposed to this problem because shared tokenizers and representational spaces provide implicit entity anchors.
Heterogeneous communication therefore creates a demanding setting in which continuous contextual transfer and exact discrete entity fidelity must be maintained simultaneously.

To address this bottleneck, we propose \textbf{\textsc{XBridge}}, a decode-free communication protocol that mitigates the entity grounding problem via dual channels. 
First, \textbf{Lexical Anchor Mapping (LAM)} deterministically translates the sender's original context tokens into the receiver's vocabulary, providing discrete entity anchors without autoregressive decoding. 
Second, a \textbf{Latent Enrichment Bridge (LEB)} uses gated cross-attention to let the receiver query the sender's hidden states, supplying continuous contextual information that can be bound to these lexical anchors. 
By decoupling the preservation of discrete entity identity (the \textit{who} and \textit{what}) from the transfer of continuous contextual knowledge (the \textit{how} and \textit{why}), \textsc{XBridge} enables heterogeneous models to communicate without requiring textual decoding or direct latent-space compatibility. 
The bridge is lightweight, requiring only 264M trainable parameters (3.8\% of the receiver), training on a small balanced sample set in under 10 minutes, and adding negligible inference overhead.

In summary, our core contributions are:
\begin{itemize}[leftmargin=*, topsep=0pt, itemsep=2pt]
    \item We formulate the entity grounding problem in heterogeneous LLM communication and demonstrate rare-token compression collapse, a failure mode in which continuous bridges preserve contextual signals but lose discrete entity identity.

    \item We propose \textsc{XBridge}, a decode-free dual-channel protocol that combines Lexical Anchor Mapping for receiver-native entity anchors with Latent Enrichment Bridge for sender-side contextual enrichment. To our knowledge, this is the first protocol to pair an explicit discrete-anchor channel with latent transfer for heterogeneous communication.

    \item We show that \textsc{XBridge} improves heterogeneous communication across three model families and seven benchmarks, achieving 11$\times$ lower latency than text-based communication, and that in a same-architecture setting it exceeds KV-cache sharing on 6 of 7 tasks.

\end{itemize}

%% file: figures/fig_hero.tex
\begin{figure*}[t]
    \centering
    \includegraphics[width=\textwidth]{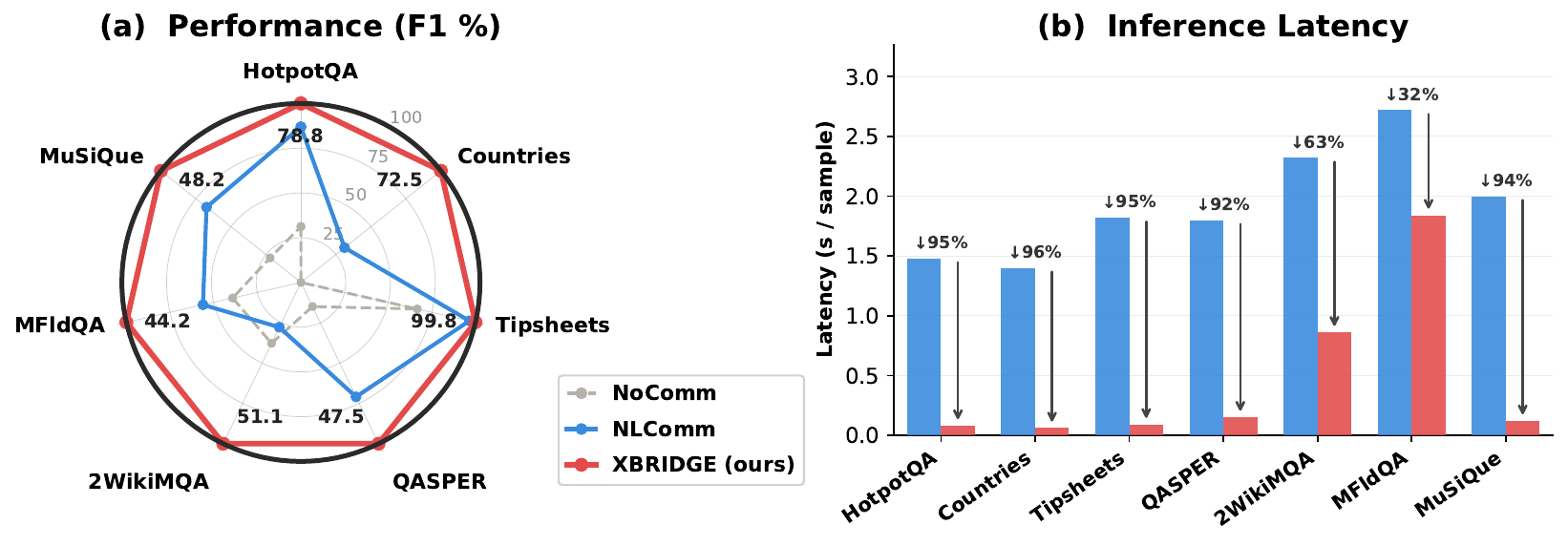}
    \caption{\textsc{XBridge} performance and 
    latency (Llama-3.1-8B $\to$ Qwen2.5-7B). 
    \textbf{(a)} F1 scores across seven 
    benchmarks. NoComm: receiver only (no communication). 
    NLComm: heterogeneous text-based communication~\citep{du2024debate}. 
    \textbf{(b)} Per-sample 
    inference latency on H200 GPU. Speedup 
    labels indicate \textsc{XBridge} vs.\ 
    NLComm.}
    \label{fig:hero}
    \vspace{-1.4em}
\end{figure*}

%% file: sections/2_motivation.tex
\section{The Entity Grounding Problem}
\label{sec:problem}
\input{figures/fidelity}

Heterogeneous communication requires satisfying three requirements.
First, the receiver must recover receiver-native lexical identities for entities that appear in the sender context.
Second, it must use contextual information encoded in the sender's hidden states.
Third, it must consume this information without treating sender activations as native receiver activations.
We refer to these requirements as \emph{entity fidelity}, \emph{entity grounding}, and \emph{representation compatibility}, respectively.

We formalize heterogeneous LLM communication as information transfer between models with incompatible tokenizers, vocabularies, and hidden spaces.
Let
\begin{equation}
    M_i = (\tau_i, V_i, E_i, F_i, G_i), \qquad i \in \{S,R\},
\end{equation}
denote a sender $M_S$ and receiver $M_R$, where $\tau_i$ is the tokenizer, $V_i$ the vocabulary, $E_i$ the embedding map, $F_i$ the transformer, and $G_i$ the language-model head.
The models are heterogeneous when
\begin{equation}
    V_S \neq V_R, \qquad d_S \neq d_R, \qquad F_S \not\equiv F_R,
\end{equation}
where architectural mismatch may include differences in depth, attention layout, normalization, positional encoding, tokenizer design, or training distribution.

The sender observes a context $C$, while the receiver observes a question $Q$ and must generate an answer $A$.
The sender performs a single forward pass:
\begin{equation}
    c_S = \tau_S(C) = (c_1,\ldots,c_{T_C}), \qquad
    H_S = F_S(E_S(c_S)) \in \mathbb{R}^{T_C \times d_S}.
\end{equation}
Unless otherwise stated, $H_S$ denotes the sender's last-layer hidden states.
A communication protocol produces a sender-side message $
    m_S = \pi_S(c_S,H_S),
$
which the receiver consumes to generate
$
    p_\pi(A \mid C,Q)
    =
    p_R(A \mid Q;\Gamma_\theta(Q,m_S)).
$
Here, $\Gamma_\theta$ denotes the receiver-side mechanism for consuming the message.
Text protocols communicate strings or token sequences, latent protocols communicate continuous states, and hybrid protocols communicate both $
    m_S =
    \bigl(
        m_{\mathrm{disc}}(C),
        m_{\mathrm{cont}}(C)
    \bigr)$.

\subsection{Entity Fidelity}
\label{sec:entity-fidelity}

Let $\mathcal{E}(C)$ denote the set of entity mentions in $C$, including names, numbers, dates, and other rare or semantically specific spans.
For an entity $e \in \mathcal{E}(C)$, define its receiver-native lexical realization as
\begin{equation}
    a_R(e) = \tau_R(e) \in V_R^\ast,
\end{equation}
where $V_R^\ast$ is the set of finite token sequences over the receiver vocabulary.
We call $a_R(e)$ an \emph{entity anchor}.
Entity fidelity is the requirement that the receiver assigns high probability to the correct receiver-native anchor.

For a gold answer entity $e^\star$, we measure entity fidelity by the receiver-side rank
\begin{equation}
    \mathrm{rank}_\pi(e^\star;C,Q)
    =
    \mathrm{rank}_{v\in V_R}
    \left[
        \ell_R(v \mid Q;\Gamma_\theta(Q,m_S))
    \right]_{v\in a_R(e^\star)},
\end{equation}
where $\ell_R$ is the receiver logit induced by $G_R$.
For multi-token entities, we use the first answer token as a diagnostic proxy.
A protocol has high entity fidelity when this rank is low.

Entity fidelity can also be characterized interventionally.
Let $C^{e\rightarrow e'}$ denote the context obtained by replacing every occurrence of $e$ with another entity $e'$ of the same semantic type.
A protocol is entity-sensitive if
\begin{equation}
    \hat{A}_\pi(C,Q)=e
    \quad\Longrightarrow\quad
    \hat{A}_\pi(C^{e\rightarrow e'},Q)=e'.
\end{equation}
This condition separates entity identity from relational structure.
If the relation is preserved but the answer does not follow the substituted entity, the protocol does not preserve entity fidelity.

\subsection{Entity Grounding Failure}
\label{sec:entity-grounding}

Entity fidelity concerns whether the receiver outputs the correct entity.
Entity grounding concerns whether the sender's contextual signal is bound to the correct receiver-native entity anchor.
A protocol may transfer contextual information while still leaving that information ungrounded to the correct entity in $V_R$.
A continuous bridge can be written abstractly as
$
    r_\theta^{(\ell)}
    =
    B_\theta^{(\ell)}(H_S,h_R^{(\ell)})
    \in
    \mathbb{R}^{T_R \times d_R},
$
where $h_R^{(\ell)} \in \mathbb{R}^{T_R \times d_R}$ is the receiver hidden-state sequence at layer $\ell$ with the receiver sequence length $T_R$.
This abstraction covers projections, adapters, hidden-state fusion, and cross-attention modules.
Such a bridge may transmit contextual information encoded in $H_S$, but by itself it does not expose receiver-native anchors $a_R(e)$.

The resulting failure is an identifiability problem.
Consider two contexts $C_e$ and $C_{e'}$ that differ primarily by replacing entity $e$ with another entity $e'$ while preserving the same relational structure.
A continuous-only protocol is vulnerable when
$
    d
    \bigl(
        m_{\mathrm{cont}}(C_e),
        m_{\mathrm{cont}}(C_{e'})
    \bigr)
    \leq
    \epsilon,
$
while the required receiver-native anchors remain distinct:
$
    a_R(e) \neq a_R(e').
$
In this regime, the receiver may recover the context but struggles to determine which discrete receiver-vocabulary entity instantiates the answer.
We refer to this as \emph{rare-token compression collapse}: continuous representations preserve contextual semantics but lose reliable access to rare or specific lexical identities.

\subsection{Representation Mismatch}
\label{sec:representation-mismatch}

Even if entity identity is preserved, sender activations cannot be directly treated as receiver activations.
Let $\mu_R^{(\ell)}$ denote the distribution of token-level hidden states encountered by receiver layer $\ell$.
For a projection $P:\mathbb{R}^{d_S}\rightarrow\mathbb{R}^{d_R}$, dimension matching only ensures that $PH_S$ lies in $\mathbb{R}^{d_R}$.
It does not imply that projected sender states match the receiver's native hidden-state distribution:
$
    PH_S \not\sim \mu_R^{(\ell)} \quad \text{in general},
$
where $\mu_R^{(\ell)}$ denotes the distribution of hidden states at the $\ell$-th layer of the receiver model.
Directly injecting $PH_S$ into the receiver residual stream or key-value cache can therefore introduce out-of-distribution activations.

A representation-compatible protocol should instead let the receiver consume sender information through its own hidden state:
\begin{equation}
    h_R^{\prime(\ell)}
    =
    h_R^{(\ell)}
    +
    g_\theta^{(\ell)}(h_R^{(\ell)},H_S),
\end{equation}
where $g_\theta^{(\ell)}$ is a learned receiver-conditioned retrieval operator.
This formulation preserves the receiver's native computation while allowing access to sender-side contextual information.

%% file: figures/fidelity.tex
\begin{figure}[t]
    \centering
    \includegraphics[width=\linewidth]{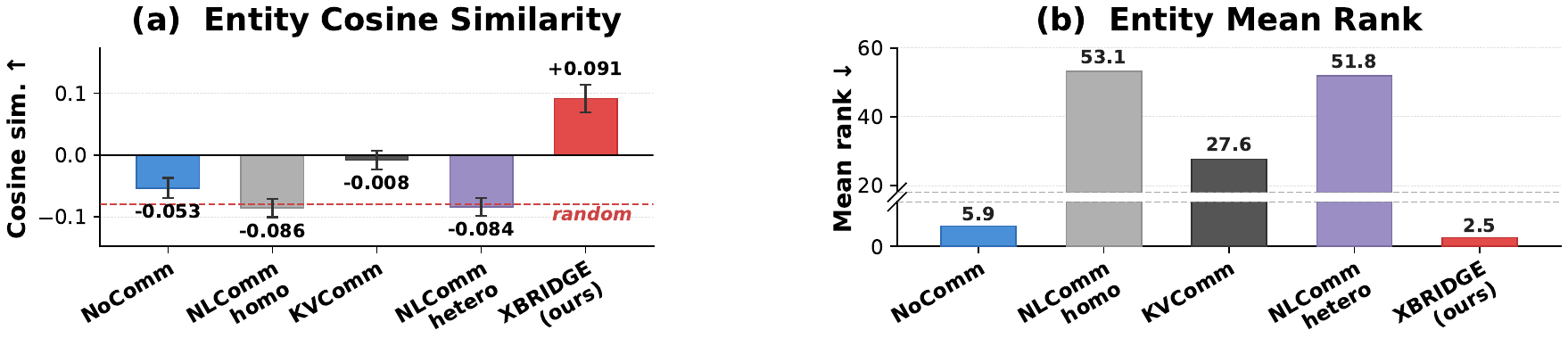}
    \caption{Entity fidelity across five communication methods
    on HotpotQA (Llama$\to$Qwen, 196 samples).
    NoComm receives no sender signal.
    NLComm \citep{du2024debate} relays text summaries
    within the same family (homo) or across families (hetero).
    KVComm \citep{shi2026kvcomm} shares KV caches within the
    same architecture.
    \textbf{(a)} Cosine similarity between generation and answer
    entity representations. Only \textsc{XBridge} achieves
    positive alignment.
    \textbf{(b)} Mean rank of the correct answer token (lower is
    better). All methods degrade rank beyond the
    no-communication baseline, while \textsc{XBridge} improves it.}
    \label{fig:motivation}
\end{figure}

%% file: sections/3_method.tex
\section{\textsc{XBridge}: Entity-Grounded Latent Communication}
\label{sec:method}
The analysis in Section~\ref{sec:problem} identifies two challenges for heterogeneous architecture communication: continuous bridges struggle to preserve discrete entity identity (Section~\ref{sec:entity-grounding}), and sender activations lie outside the receiver's representation space (Section~\ref{sec:representation-mismatch}). 
We therefore decompose heterogeneous communication into two complementary mechanisms: \textbf{Lexical Anchor Mapping} (LAM, Section~\ref{sec:LAM}) provides discrete entity anchors, and a \textbf{Latent Enrichment Bridge} (LEB, Section~\ref{sec:LEB}) provides contextual enrichment. 
Both signals are produced by a single sender forward pass with no autoregressive decoding; sender and receiver remain frozen throughout.
\input{figures/overview}

\subsection{\textsc{XBridge} Overview}
\label{sec:setup}

We adopt the asymmetric communication setting formalized in Section~\ref{sec:problem}, following the protocol established by \citet{shi2026kvcomm}, in which the sender observes a context and the receiver observes a question: the sender $M_S$ processes a context document $C$ through a single forward pass, producing last-layer hidden states $H_S \in \mathbb{R}^{T_C \times d_S}$ and context token sequence $c_S = (c_1, \dots, c_{T_C})$. 
The receiver $M_R$ holds only the question $Q$ and must generate an answer $A$ from whatever signal the sender communicates. 
\textsc{XBridge} transmits both $H_S$ and $c_S$: LAM (Section~\ref{sec:LAM}) uses $c_S$, and LEB (Section~\ref{sec:LEB}) uses $H_S$.

\subsection{Lexical Anchor Mapping (LAM) for Discrete Grounding}
\label{sec:LAM}

LAM addresses the entity grounding failure (Section~\ref{sec:entity-grounding}) by placing the sender's original context tokens, including entity mentions, as receiver-native lexical inputs.
These lexical anchors provide the discrete basis that LEB requires to ground its contextual enrichment to specific entities (Section~\ref{sec:integration}).

\paragraph{Cross-vocabulary mapping.}
Given the sender's context tokens $c_S = (c_1, \dots, c_{T_C})$ obtained by tokenizing $C$ with $M_S$'s tokenizer, each token $c_i \in \mathcal{V}_S$ is mapped to the receiver's vocabulary $\mathcal{V}_R$ via a deterministic, training-free mapping function $\phi: \mathcal{V}_S \to \mathcal{V}_R^{*}$, precomputed offline once per sender--receiver tokenizer pair. 
For tokens whose surface string appears in both vocabularies, $\phi$ performs a direct ID-to-ID lookup. 
For the remainder, a string fallback decodes the token to its surface string and re-tokenizes it with the receiver's tokenizer, producing one or more receiver tokens expanded in place. 
The mapping is lossless and adds negligible latency ({$<$}1\,ms); detailed statistics and examples are in Appendix~\ref{appendix:vocab-mapping}.

\paragraph{Receiver-native lexical embeddings.}
The mapped token IDs are converted to receiver embeddings $e_\text{ctx} = (E_R[\phi(c_1)], \dots, E_R[\phi(c_{T_C})])$ via the receiver's frozen embedding matrix $E_R$, omitting expansion from re-tokenization for clarity. 
These embeddings are prepended to the receiver's question input, following the input prepend convention used in soft-token communication \citep{pham2024cipher}:
\begin{equation}
    x_R = [e_\text{ctx} \;;\; E_R(Q)]
\end{equation}
By using $E_R$ as a lookup table rather than training a separate projection, transmitted tokens are guaranteed to lie in the same space as the receiver's own input tokens, ensuring that pretrained self-attention patterns apply directly to the transferred context. Unlike prior methods that require the sender to autoregressively decode a summary \citep{du2024debate, pham2024cipher}, LAM transfers the full context without decoding, eliminating both latency and information loss from fixed-length compression.

\subsection{Latent Enrichment Bridge (LEB) for Context Adaptation} 
\label{sec:LEB}

LAM preserves entity identity but transmits only the sender's raw tokens, not the sender's \textit{contextual understanding} of those tokens. 
The sender's last-layer hidden states $H_S$ encode information integrated across all context positions through the sender's full transformer stack, information that raw token embeddings cannot convey. 
We use the last layer because it produces the most globally-integrated representations, as confirmed by a layer sweep in Appendix~\ref{appendix:arch-ablation}. However, $H_S$ lies in the sender's representation space, which the receiver has never encountered during training (Section~\ref{sec:representation-mismatch}).
Directly injecting these representations, whether by prepending hidden states or replacing KV caches, places out-of-distribution vectors into the receiver's processing pipeline and degrades performance.
\textit{Cross-attention} provides a natural solution: rather than injecting sender representations into the receiver, the receiver queries the sender in its own native space. 
This receiver-driven mechanism has three stages, illustrated in the bridge panel of Figure~\ref{fig:method-overview}.

\paragraph{Projections.}
We insert $M{=}4$ gated cross-attention modules into the receiver, one after each of layers $\ell \in \{7, 14, 21, 28\}$ of the receiver's 28 layers, following the insertion density of Flamingo~\citep{alayrac2022flamingo}.
Keeping the backbone frozen and training only lightweight inserted modules follows the parameter-efficient adaptation practice used across domains~\citep{lee2024subgraph}.
We find four modules optimal; fewer provide insufficient capacity while more overfit with limited training data (Appendix~\ref{appendix:arch-ablation}).
Each module has its own learnable parameters (${\sim}$66M each, ${\sim}$264M total, 3.8\% of the receiver). Both models remain frozen; only the bridge modules are trained.
At each insertion layer $\ell$, the receiver's hidden state $h_R^{(\ell)} \in \mathbb{R}^{d_R}$ is projected into a query, while the sender's hidden states $H_S$ are projected into keys and values:
\begin{align}
    Q^{(\ell)} &= W_Q^{(\ell)} \cdot \mathrm{LN}_R^{(\ell)}(h_R^{(\ell)}), \quad
    K^{(\ell)} = W_K^{(\ell)} \cdot \mathrm{LN}_S^{(\ell)}(H_S), \quad
    V^{(\ell)} = W_V^{(\ell)} \cdot \mathrm{LN}_S^{(\ell)}(H_S)
\end{align}
where $W_Q^{(\ell)} \in \mathbb{R}^{d_R \times d_R}$, $W_K^{(\ell)}, W_V^{(\ell)} \in \mathbb{R}^{d_R \times d_S}$, and $\mathrm{LN}_R^{(\ell)}$, $\mathrm{LN}_S^{(\ell)}$ are separate layer normalizations for the receiver and sender. 
The projection matrices simultaneously handle the dimensional mismatch ($d_S \to d_R$), serving as the learned translation between representation spaces. 
All four modules query the same $H_S$, but each has its own parameters, enabling different layers to extract different aspects of the sender's representation.

\paragraph{Cross-attention.}
The receiver's query selects relevant sender positions through scaled dot-product attention:
\begin{equation}
    A^{(\ell)} = \mathrm{softmax}\!\left(Q^{(\ell)} K^{(\ell)\top} / \sqrt{d_k}\right) \cdot V^{(\ell)}
\end{equation}
This is the key advantage over injection-based alternatives: the receiver decides what information to retrieve based on its current processing state at layer $\ell$, rather than receiving a fixed representation regardless of what it needs.

\paragraph{Gated residual.}
The cross-attention output is added to the receiver's hidden state via a gated residual connection, preserving the receiver's pretrained representations while incorporating bridge information:
\begin{equation}
    h_R'^{(\ell)} = h_R^{(\ell)} + \tanh(\alpha^{(\ell)}) \cdot A^{(\ell)}
    \label{eq:gated-residual}
\end{equation}
where $h_R'^{(\ell)}$ is the updated hidden state that continues through the receiver's subsequent layers. 
We initialize $\alpha^{(\ell)} = 1.0$ ($\tanh(1.0) \approx 0.76$) so that the bridge contributes immediately, unlike Flamingo's zero initialization which requires the model to discover the bridge signal during training. 
With our limited training budget, warm initialization converges faster (Appendix~\ref{appendix:arch-ablation}).

The bridge is trained with standard next-token prediction loss on the answer tokens:
\begin{equation}
    \mathcal{L}(\theta) = - \sum_{t=1}^{|A|} 
    \log P_R(a_t \mid a_{<t}, e_\text{ctx}, Q; 
    \theta_\text{bridge})
\end{equation}
where $e_\text{ctx}$ is computed via LAM (Section~\ref{sec:LAM}). Since the sender is frozen, its outputs ($H_S$ and context token IDs) are precomputed and cached, reducing training to under 10 minutes on a single GPU on a balanced set of 587 samples drawn across the seven task domains. 
Once trained, the bridge is fixed and applied to all tasks at inference time without further adaptation (Section~\ref{sec:experiments}). 
Training data composition is analyzed in Appendix~\ref{appendix:data-composition}.
The bridge adapts to different model pairs by adjusting only the projection dimensions; we train separate instances for each pair using the same protocol.

\subsection{Entity-Grounded Bridge Integration}
\label{sec:integration}

The two mechanisms are integrated through the receiver's own processing, without any explicit fusion layer. 
LAM places entity mentions (e.g., ``Christopher Nolan'') in the receiver's input as discrete embeddings. 
As these propagate through the receiver's layers via self-attention, they shape the receiver's hidden states $h_R^{(\ell)}$ at every depth, embedding entity information into the representations that LEB modules use as queries. 
This is the grounding mechanism: when LEB at layer $\ell$ computes $Q^{(\ell)} = W_Q^{(\ell)} \cdot \mathrm{LN}_R^{(\ell)}(h_R^{(\ell)})$, the query already encodes which entities are present and where they appear, enabling the cross-attention to focus on the sender positions that carry relevant contextual information about those entities. 
Without LAM, $h_R^{(\ell)}$ reflects only the question, producing queries too vague to extract entity-specific signals from the sender, so the LEB conveys contextual signals while entity identity is lost in the continuous bottleneck, which is why LEB-only performance drops to $\sim$30\% F1 (Section~\ref{sec:ablation}).
Without LEB, the receiver has the sender's context tokens but lacks the sender's contextual understanding of their relationships. 
Together, LEB's contextual enrichment is anchored to specific entities rather than floating in continuous space. 
We verify this complementarity quantitatively in Section~\ref{sec:ablation}.

%% file: figures/overview.tex
\begin{figure}[t]
    \centering
    \includegraphics[width=\linewidth]{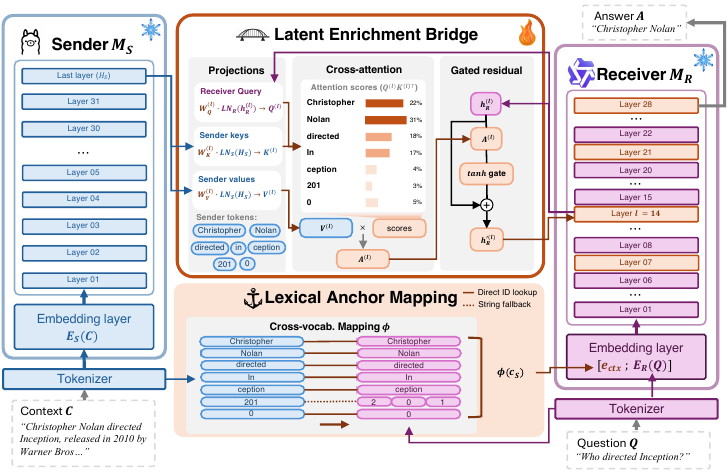}
    \caption{\textsc{XBridge} overview. A single sender forward pass produces hidden states $H_S$ and context tokens $c_S$. LAM maps $c_S$ to receiver embeddings (entity grounding). LEB lets the receiver query $H_S$ at four layers (contextual enrichment). The receiver generates the answer conditioned on both signals.}
    \label{fig:method-overview}
\end{figure}

%% file: sections/4_experiments.tex
\section{Experiments}
\label{sec:experiments}
\input{tables/tab_hetero_results}
\subsection{Experimental Setup}
\label{sec:exp-setup}

\paragraph{Models.}
We evaluate communication across three model families to test cross-architecture generality: Llama-3.1-8B-Instruct \citep{dubey2024llama3} 
(Llama), Qwen2.5-7B-Instruct \citep{qwen2024qwen25} (Qwen), and 
Mistral-7B-Instruct-v0.3 (Mistral). 
These families differ in architecture (hidden dimension, layer count, 
attention structure), tokenizer design, and training data, ensuring that improvements are not an artifact of a single architecture pairing. 
We evaluate three heterogeneous pairs: Llama$\to$Qwen, Qwen$\to$Llama, and Mistral$\to$Qwen. 
We also verify that the framework generalizes to same-architecture settings (Appendix~\ref{appendix:homo}). 
To test whether larger senders produce richer representations for LEB, we additionally evaluate sender size scaling with Qwen2.5-\{1.5B, 7B, 14B\}-Instruct$\to$Llama.

\paragraph{Tasks and protocol.}
We evaluate on seven benchmarks: HotpotQA \citep{yang2018hotpotqa}, MuSiQue \citep{trivedi2022musique}, QASPER \citep{dasigi2021qasper}, 2WikiMultihopQA (2Wiki) and MultiFieldQA-en (MFldQA) from LongBench \citep{bai2024longbench}, Countries and Tipsheets \citep{shi2026kvcomm}. 
The sender receives the context document only and the receiver receives the question only, the asymmetric protocol established by \citet{shi2026kvcomm}, which we apply identically to every method compared here.
For NLComm baselines, the sender generates a 128-token summary via greedy decoding. 
We report F1 score. 
Task details are in Appendix~\ref{appendix:tasks}.

\paragraph{Compared methods.}
We compare against natural language communication (NLComm) \citep{du2024debate}, where the sender generates a text summary and relays it to the receiver. 
We distinguish NLComm\textsubscript{hetero} (cross-family) and NLComm\textsubscript{homo} (same family).
We also report two single-model reference points measured with the receiver alone. NoComm, where the receiver sees only the question, is a lower reference indicating what the receiver achieves without communication. FullComm, where the receiver reads the full context directly, removes the communication problem rather than solving it, and serves as an upper reference in the sense of the Skyline used by \citet{shi2026kvcomm}. Neither is a competing communication method, and both are excluded from the best-per-task comparison.
Same-architecture methods, including NLComm\textsubscript{homo} and KVComm \citep{shi2026kvcomm} which shares KV caches between models of the same family, are compared in Appendix~\ref{appendix:homo}.

\subsection{Main Results}
\label{sec:main-results}

Table~\ref{tab:main-results} presents heterogeneous communication results across all three model pairs and seven benchmarks. \textsc{XBridge} outperforms NLComm\textsubscript{hetero} on all 7 tasks across all three model pairs, with average gains of +21.4\,pp (Llama$\to$Qwen), +14.3\,pp (Qwen$\to$Llama), and +20.9\,pp 
(Mistral$\to$Qwen). 
The improvements are largest on tasks requiring multi-hop reasoning or world knowledge (Countries, 2WikiMQA, MuSiQue), where LEB provides contextual enrichment that text summaries cannot convey.
On average, \textsc{XBridge} surpasses even FullComm across all three model pairs (63.2 vs. 55.2 for Llama$\rightarrow$Qwen, 61.9 vs. 61.2 for Qwen$\rightarrow$Llama, 65.0 vs. 55.2 for Mistral$\rightarrow$Qwen). Since LAM places the full mapped context in the receiver's input, \textsc{XBridge} sees what FullComm sees and adds the sender's hidden states through LEB, so the comparison is not one of indirect against direct access. The same-architecture setting isolates this: with Qwen2.5-7B sending to itself, LAM reduces to an identity mapping, so the receiver's input matches FullComm's and only LEB differs, and \textsc{XBridge}$_{\text{homo}}$ still leads by 7.9 pp on average (Appendix~\ref{appendix:homo}). With identical models and identical input, the gain comes from LEB carrying the sender's integrated representation that the receiver would otherwise recompute.
The optimal sender varies by task: Mistral$\to$Qwen achieves the highest scores on HotpotQA, QASPER, and MuSiQue, while Llama$\to$Qwen leads on MFldQA and 2WikiMQA. 
This complementarity suggests that selecting the sender based on task characteristics can further improve performance. 
In the same-architecture setting, \textsc{XBridge} also exceeds KVComm \citep{shi2026kvcomm} on 6 of 7 tasks 
(Appendix~\ref{appendix:homo}).

\subsection{Ablation: Both Anchors and Enrichment Are Necessary}
\label{sec:ablation}
\input{tables/tab_perturbation}
The gains reported above rely on both LAM and LEB working together. 
To understand how much each contributes, we remove them individually on HotpotQA (Table~\ref{tab:grounding-ablation}).
\textsc{XBridge} without LAM (\textsc{XBridge}$_\text{-LAM}$) drops to 30.3\% F1, close to NoComm: LEB receives continuous signals but without entity anchors these signals are not bound to specific entities, confirming the entity grounding failure identified in Section~\ref{sec:entity-grounding}. 
\textsc{XBridge} without LEB (\textsc{XBridge}$_\text{-LEB}$) reaches 56.5\%, providing entity grounding but no contextual enrichment beyond what the tokens themselves contain. 
The full \textsc{XBridge} achieves 78.8\%, exceeding the FullComm. 
The +22.3\,pp gain from adding LEB to LAM contrasts with the modest +5.7\,pp from LEB-only over NoComm, demonstrating that grounding is the enabling condition for LEB effectiveness.
LEB contribution scales with task reasoning complexity: from +35.8\,pp on factual inference requiring world knowledge (Countries) to +2.0\,pp on structured extraction (Tipsheets), with per-task breakdown in Appendix~\ref{appendix:bridge-contribution}.

\subsection{Bridge Mechanism and Role Separation}
\label{sec:mechanism-role}
The ablation confirms that both mechanisms are necessary. A deeper question follows: does each carry a distinct type of information, or do they redundantly encode the same signal? 
To test this, we run an entity perturbation experiment on 100 HotpotQA samples, replacing every occurrence of the answer entity in the context with an unrelated entity of the same type (e.g., ``Christopher Nolan'' $\to$ ``Bong Joon Ho''), then independently control which version each mechanism receives (Table~\ref{tab:perturbation}).

Swapping LEB alone does not change the output entity: conditions A and D produce identical outputs (45 original, 0 swapped, 55 neither). 
The ``neither'' category contains outputs that mention neither the original nor the substituted entity, such as a date, another entity, or an unrelated span; we therefore interpret the experiment by comparing the original-versus-swapped direction among target-entity outputs.
Under this interpretation, swapping LAM shifts the output toward the substituted entity regardless of which LEB version is used; conditions B and C produce nearly identical counts (37 swapped, 4--5 original, 58--59 neither).
These results indicate a clear role separation: LAM supplies \textit{which} entities appear in the output, while LEB supplies \textit{how} the receiver reasons about them.

\input{figures/5cond}
Figure~\ref{fig:mechanism-5cond} examines LEB's effect on the receiver's internal state across five communication conditions on 196 HotpotQA samples.
NLComm methods attend most heavily to answer positions (2.0--2.7$\times$ above average) yet achieve the worst token rank because paraphrased text replaces exact entity mentions.
\textsc{XBridge} is the only method that produces positive cosine alignment between generation and answer entity representations, translating to correct-token probability of 0.752 vs.\ 0.635--0.662 for all other conditions (94\% of individual samples, Appendix~\ref{appendix:scatter}).

\input{tables/tab_ablation_size_inference}
\subsection{Beyond Accuracy: Latency, Composability, and Sender Scaling}
\label{sec:beyond-accuracy}

\textsc{XBridge} achieves 11$\times$ lower latency than NLComm (0.15\,s vs.\ 1.70\,s per sample on HotpotQA) by eliminating autoregressive decoding: the sender performs a single forward pass (${\sim}$0.05\,s) with negligible bridge overhead ({$<$}0.01\,s), as reported in Table~\ref{tab:latency}. 
The modular bridge design also enables zero-shot composability: two independently trained bridges (Llama and Mistral as senders) combine at inference without retraining, achieving 70.4\% F1 on HotpotQA compared to 67.0\% for single-sender full context and 56.8\% for dual NLComm (Appendix~\ref{appendix:multiagent}). 
Finally, LAM is sender-size invariant since it depends only on the vocabulary mapping, not the sender's capacity: even a 1.5B sender achieves 59.5\%, approaching FullComm, while the 7B and 14B senders surpass it (Table~\ref{tab:sender-scaling}).

%% file: tables/tab_hetero_results.tex
\begin{table*}[t]
\centering
\caption{Heterogeneous communication results 
(F1\,\%). The best score per task within each model 
pair is bolded and the Improv.\ row shows the gain 
of \textsc{XBridge} over the second-best method, both 
excluding FullComm.}
\label{tab:main-results}
\small
\newcolumntype{C}[1]{>{\centering\arraybackslash}p{#1}}
\renewcommand{\arraystretch}{0.9}
\begin{tabular}{@{}l C{1.0cm} C{1.0cm} C{1.0cm} C{1.0cm} C{1.0cm} C{1.0cm} C{1.0cm} C{0.7cm}@{}}
\toprule
\textbf{Method} & \textbf{Countries} & \textbf{Tipsheets} & \textbf{HotpotQA} & \textbf{QASPER} & \textbf{MuSiQue} & \textbf{MFldQA} & \textbf{2Wiki} & \textbf{Avg} \\
\midrule
\multicolumn{9}{c}{\textit{$M_S$: Llama-3.1-8B-Instruct; $M_R$: Qwen2.5-7B-Instruct}} \\
\midrule
\textbf{NoComm} & 0.0 & 66.5 & 24.6 & 7.1 & 10.6 & 17.3 & 19.3 & 20.8 \\
\textbf{FullComm} & 50.8 & 98.5 & 78.4 & 31.3 & 31.5 & 48.2 & 47.6 & 55.2 \\
\textbf{NLComm\textsubscript{hetero}} & 22.6 & 96.3 & 68.5 & 33.7 & 32.5 & 24.8 & 14.2 & 41.8 \\
\rowcolor{gray!15}
\textbf{\textsc{XBridge}} & \textbf{72.5} & \textbf{99.8} & \textbf{78.8} & \textbf{47.5} & \textbf{48.2} & \textbf{44.2} & \textbf{51.1} & \textbf{63.2} \\%
\cdashline{1-9}\\[-8pt]
\quad Improv. & \small +49.9 & \small +3.5 & \small +10.3 & \small +13.8 & \small +15.7 & \small +19.4 & \small +36.9 & \small +21.4 \\
\midrule
\multicolumn{9}{c}{\textit{$M_S$: Qwen2.5-7B-Instruct; $M_R$: Llama-3.1-8B-Instruct}} \\
\midrule
\textbf{NoComm} & 0.0 & 11.0 & 31.4 & 3.8 & 2.6 & 11.2 & 12.9 & 10.4 \\
\textbf{FullComm} & 67.0 & 95.0 & 82.9 & 36.7 & 41.3 & 53.3 & 51.8 & 61.2 \\
\textbf{NLComm\textsubscript{hetero}} & 56.2 & 94.8 & 76.8 & 34.1 & 33.5 & 21.7 & 16.3 & 47.6 \\
\rowcolor{gray!15}
\textbf{\textsc{XBridge}} & \textbf{62.0} & \textbf{99.0} & \textbf{80.6} & \textbf{46.6} & \textbf{62.8} & \textbf{40.3} & \textbf{42.1} & \textbf{61.9} \\%
\cdashline{1-9}\\[-8pt]
\quad Improv. & \small +5.8 & \small +4.2 & \small +3.8 & \small +12.5 & \small +29.3 & \small +18.6 & \small +25.8 & \small +14.3 \\
\midrule
\multicolumn{9}{c}{\textit{$M_S$: Mistral-7B-Instruct-v0.3; $M_R$: Qwen2.5-7B-Instruct}} \\
\midrule
\textbf{NoComm} & 0.0 & 66.5 & 24.6 & 7.1 & 10.6 & 17.3 & 19.3 & 20.8 \\
\textbf{FullComm} & 50.8 & 98.5 & 78.4 & 31.3 & 31.5 & 48.2 & 47.6 & 55.2 \\
\textbf{NLComm\textsubscript{hetero}} & 43.6 & 96.0 & 66.1 & 31.8 & 28.3 & 27.0 & 15.9 & 44.1 \\
\rowcolor{gray!15}
\textbf{\textsc{XBridge}} & \textbf{70.0} & \textbf{100.0} & \textbf{84.9} & \textbf{50.6} & \textbf{65.0} & \textbf{35.8} & \textbf{48.7} & \textbf{65.0} \\%
\cdashline{1-9}\\[-8pt]
\quad Improv. & \small +26.4 & \small +4.0 & \small +18.8 & \small +18.8 & \small +36.7 & \small +8.8 & \small +32.8 & \small +20.9 \\
\bottomrule
\end{tabular}
\end{table*}

%% file: tables/tab_perturbation.tex
\begin{table}[t]
\centering
\caption{Entity perturbation experiment on 100 HotpotQA samples.
Each sample's answer entity is replaced with an unrelated entity
(e.g., ``Christopher Nolan'' $\to$ ``Bong Joon Ho'').
LAM and LEB inputs are independently swapped to test which
mechanism determines entity identity. Output is classified by
which entity appears in the generated answer.}
\label{tab:perturbation}
\small
\renewcommand{\arraystretch}{0.9}
\begin{tabular}{@{}lllccc@{}}
\toprule
\textbf{Condition} & \textbf{LAM} & \textbf{LEB} & \textbf{$\to$ original} & \textbf{$\to$ swapped} & \textbf{$\to$ neither} \\
\midrule
A: Original          & original & original & 45 & 0  & 55 \\
B: Both swapped      & swapped  & swapped  & 4  & 37 & 59 \\
C: LAM swapped only  & swapped  & original & 5  & 37 & 58 \\
D: LEB swapped only  & original & swapped  & 45 & 0  & 55 \\
\bottomrule
\end{tabular}
\end{table}

%% file: figures/5cond.tex
\begin{figure*}[t]
    \centering
    \includegraphics[width=\linewidth]{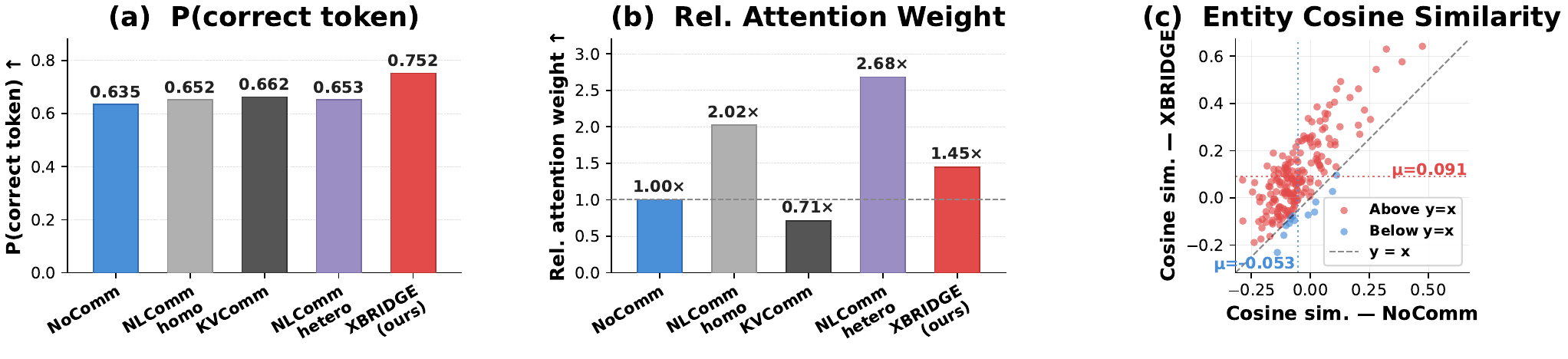}
    \caption{Bridge mechanism analysis on HotpotQA
    (Llama$\to$Qwen, 196 samples).
    \textbf{(a)} Probability assigned to the correct answer token.
    \textbf{(b)} Relative attention weight on answer positions.
    NLComm attends most heavily yet predicts worst (cf.\ Figure 2b),
    revealing an attention-prediction disconnect.
    \textbf{(c)} Per-sample entity cosine similarity comparing
    NoComm and \textsc{XBridge}. Points above $y{=}x$ indicate
    bridge improvement.}
    \label{fig:mechanism-5cond}
\end{figure*}

%% file: tables/tab_ablation_size_inference.tex
\begin{table}[t]
\begin{minipage}[t]{0.28\linewidth}
\centering
\caption{Entity grounding ablation on HotpotQA, 
Llama$\to$Qwen (F1\,\%).}
\label{tab:grounding-ablation}
\small
\renewcommand{\arraystretch}{0.9}
\begin{tabular}{@{}lcc@{}}
\toprule
\textbf{Config.} & \textbf{F1} & \textbf{$\Delta$} \\
\midrule
\textbf{NoComm} & 24.6 & -- \\
\textbf{\textsc{XBridge\textsubscript{-LAM}}}& 30.3 & +5.7 \\
\textbf{\textsc{XBridge\textsubscript{-LEB}}} & 56.5 & +31.9 \\
\textbf{\textsc{XBridge}} & \textbf{78.8} & +54.2 \\
\bottomrule
\end{tabular}
\end{minipage}%
\hfill
\begin{minipage}[t]{0.26\linewidth}
\centering
\caption{Sender size scaling: Qwen $\to$ Llama, \textsc{XBridge} (F1\,\%, 7-task avg).}
\label{tab:sender-scaling}
\small
\renewcommand{\arraystretch}{0.9}
\begin{tabular}{@{}lc@{}}
\toprule
\textbf{Sender} & \textbf{Avg} \\
\midrule
\textbf{Qwen2.5-1.5B} & 59.5 \\
\textbf{Qwen2.5-7B} & 61.9 \\
\textbf{Qwen2.5-14B} & \textbf{62.1} \\
\bottomrule
\end{tabular}
\end{minipage}%
\hfill
\begin{minipage}[t]{0.42\linewidth}
\centering
\caption{Inference latency per sample on HotpotQA.}
\label{tab:latency}
\small
\renewcommand{\arraystretch}{0.9}
\begin{tabular}{@{}lcc@{}}
\toprule
\textbf{Method} & \textbf{s/sample} & \textbf{Rel. Latency} \\
\midrule
\textbf{NoComm} & 0.09 & 1.0$\times$ \\
\textbf{FullComm} & 0.13 & 1.4$\times$ \\
\textbf{NLComm\textsubscript{homo}} & 1.70 & 18.9$\times$ \\
\textbf{NLComm\textsubscript{hetero}} & 1.70 & 18.9$\times$ \\
\textbf{KVComm} & 0.13 & 1.4$\times$ \\
\textbf{\textsc{XBridge}} & \textbf{0.15} & \textbf{1.7$\times$} \\
\bottomrule
\end{tabular}
\end{minipage}
\end{table}

%% file: sections/7_conclusion.tex
\section{Conclusion}
\label{sec:conclusion}

In this work, we identified the entity grounding problem in heterogeneous LLM communication, where continuous bridges lose entity identity in the latent bottleneck. 
We proposed \textsc{XBridge}, a decode-free protocol that addresses this by combining LAM with LEB, decomposing communication into discrete entity anchors and continuous contextual enrichment. 
Extensive experiments across three model families, seven benchmarks, and both communication directions demonstrated that \textsc{XBridge} outperforms NLComm on every task with 11$\times$ lower latency. 
We highlight the complementarity of the two mechanisms: LAM determines which entities appear in the output, while LEB determines how the receiver reasons about them, as confirmed by the entity perturbation experiment. 
Our work shows that heterogeneous LLMs can communicate at the latent level while preserving entity fidelity, and we see this as an early step toward cross-architecture multi-agent collaboration.

%% file: sections/limitation.tex
\section*{Limitations}
\label{sec:limitations}

XBridge is trained for a single sender to receiver direction, so one bridge carries information one way over one exchange. A two-way setup would add a second bridge for the reverse direction, and since each bridge is lightweight and decode-free, the cost per exchange would stay low, but multi-turn latent dialogue, where the two models alternate and each turn conditions on what the other sent, is beyond what we evaluate here. The same holds for the number of agents: we have tested composition of two independently trained bridges, leaving open how bridge signals interact when more agents participate. Extending the protocol along both directions, bidirectional exchange and many-agent composition, is what we see as the path toward a cross-architecture communication protocol suited to full multi-agent systems.

%% file: sections/x_appendix.tex
\appendix

\section{LAM Details}
\label{appendix:lam-details}

\subsection{Cross-Vocabulary Mapping}
\label{appendix:vocab-mapping}

Table~\ref{tab:vocab-stats} summarizes the vocabulary overlap and
token transfer statistics for the two sender--receiver pairs used
in our experiments.

\input{tables/tab_cross-vocab}

Table~\ref{tab:vocab-examples} shows representative examples of
tokenization mismatches and how the string-based fallback resolves
them.

\input{tables/tab_vocab_example}

The mismatch patterns reflect fundamental differences in tokenizer
design. Llama-3.1 uses a 128K BPE vocabulary that overlaps
substantially with Qwen2.5's 152K vocabulary, resulting in
mismatches confined almost entirely to multi-digit numbers.
Mistral-7B uses a smaller 32K SentencePiece vocabulary, causing
common English words and proper nouns to fall back to string-based
re-tokenization. In both cases, the fallback is lossless: the
surface string content is fully preserved, with only positional
expansion as a side effect.

\subsection{Grounding Format Independence}
\label{appendix:format-independence}

A critical question is whether LEB's effectiveness depends on the specific grounding format (embedding prepend vs.\ text prompt) or merely on the \textit{presence} of entity anchors.

\input{tables/tab_format_independence}

LEB consistently improves all grounding formats ($+12.5$ to $+22.3$\,pp), demonstrating that entity presence, not format, is what enables LEB. Whether entity anchors arrive as prepended embeddings, as text within a chat template, or as argmax-decoded tokens, LEB extracts contextual enrichment from the sender's $H_S$ via cross-attention regardless of how those anchors enter the receiver.

The smaller $\Delta$ for NLComm ($+12.5$\,pp) reflects its higher base: NLComm text already contains sender reasoning, leaving less room for LEB improvement. Notably, NLComm with LEB (81.0\%) surpasses LAM with LEB (78.8\%), because the sender's autoregressive decoding performs reasoning compression that benefits the receiver. However, LAM remains the preferred grounding format because it eliminates decoding latency entirely ($14\times$ faster than NLComm) while achieving comparable performance with LEB.

\section{LEB Design}
\label{appendix:leb-design}

\subsection{Bridge Architecture Ablations}
\label{appendix:arch-ablation}

We ablate bridge design decisions on HotpotQA (Llama$\to$Qwen). All variants use balanced training (587 samples).

\input{tables/tab_arch_ablation}

\paragraph{Number of bridge modules.}
We vary the number of cross-attention modules inserted into the receiver (Qwen2.5-7B, 28 layers). 4 modules (264M parameters) is optimal. 7 modules (462M) overfits due to the higher parameter-to-data ratio with balanced training. 2 modules (132M) provides insufficient capacity for cross-architecture translation. This follows Flamingo's \citep{alayrac2022flamingo} finding that sparser insertion is preferred for larger models.

\paragraph{Sender layer selection.}
Bridge performance increases monotonically from early to final sender layers, with layers before the midpoint falling below the no-bridge baseline. The last layer is optimal because it integrates information across all context positions through the sender's full attention stack, producing the most globally-aware representations.

\paragraph{Gate initialization.}
We initialize the $\tanh$ gate at $\alpha=1.0$ ($\tanh(1.0) \approx 0.76$), differing from Flamingo's zero initialization. With $\alpha=0$, the bridge is initially invisible to the receiver, requiring training to ``discover'' the bridge signal. With $\alpha=1.0$, the bridge immediately contributes cross-architecture context. In our setting with limited training data (587 samples), warm initialization converges faster and achieves higher final performance.

\paragraph{Auxiliary losses.}
Adding a contrastive loss (encouraging the bridge to distinguish correct sender context from random contexts) does not improve performance beyond NTP loss alone. The NTP objective already implicitly requires the bridge to extract useful information from the sender's hidden states.

\subsection{Bridge Training Data Composition}
\label{appendix:data-composition}

The composition of bridge training data significantly affects per-task performance.

\input{tables/tab_data_composition}

\textit{HotpotQA-only bridge} (20K samples) is inferior even on its own training domain (76.1\% vs.\ 78.8\% for the balanced bridge), and catastrophically degrades on unseen task types: Countries drops to 27.7\% and QASPER to 36.2\%. Overfitting to a single task's context-length distribution and reasoning pattern prevents the bridge from generalizing.

\textit{Unbalanced universal bridge} (42K samples, 94\% HotpotQA+MuSiQue) improves over the single-task bridge across all tasks but remains below the balanced bridge on average (59.3\% vs.\ 63.2\%). The heavy skew toward multi-hop QA patterns limits generalization to dissimilar tasks such as Countries and QASPER.

\textit{Balanced universal bridge} (587 samples, $\sim$100 per task) achieves the best performance on every metric despite using $30\times$ less training data than the unbalanced variant. The balanced bridge learns task-appropriate gating: it remains near-closed for near-saturated factual tasks (Tipsheets 99.8\%) while actively contributing on complex reasoning tasks (HotpotQA 78.8\%, Countries 72.5\%). The average across all 7 tasks (63.2\%) is the highest among all bridge variants.

\paragraph{Implication for sender size scaling.}
The balanced training set (587 samples) is sufficient for 7B senders but becomes a bottleneck for larger models. The 14B sender (hidden dim 5120) produces richer representations that require more training data to learn effective projections. Within this data budget, the 14B bridge plateaus at 62.1\%, matching the 7B sender (61.9\%) despite its larger capacity. We expect that scaling the balanced training set would unlock the 14B sender's advantage, but leave this to future work.

\section{Additional Results}
\label{appendix:additional-results}

\subsection{Homogeneous Comparison}
\label{appendix:homo}

To verify that \textsc{XBridge} generalizes beyond heterogeneous settings, we train a same-architecture bridge using Qwen2.5-7B as both sender and receiver. LAM reduces to an identity mapping in this setting, so the comparison isolates LEB's contribution on top of full context reading.

\input{tables/tab_homo_results}

Table~\ref{tab:homo-results} compares \textsc{XBridge}$_\text{homo}$ against NLComm\textsubscript{homo} and KVComm \citep{shi2026kvcomm}. \textsc{XBridge}$_\text{homo}$ outperforms NLComm\textsubscript{homo} on all 7 tasks and KVComm on 6 of 7 tasks, with an average improvement of +12.0\,pp over the stronger baseline. Notably, \textsc{XBridge}$_\text{homo}$ surpasses even FullComm on 6 of 7 tasks (+7.9\,pp on average), demonstrating that LEB adds value beyond what the receiver extracts from reading the context alone. The sole exception is MultiFieldQA, consistent with the pattern observed in heterogeneous settings (Section~\ref{sec:ablation}).

\subsection{Bridge Contribution per Task}
\label{appendix:bridge-contribution}

Table~\ref{tab:bridge-delta} breaks down LEB's 
contribution across all seven tasks for the 
Llama$\to$Qwen pair. LEB contribution scales 
with the reasoning complexity of the task.

\input{tables/bridge_contribution}

On Countries, the context states a person's location 
at a landmark (e.g., ``Eve is at the Uppland Runic 
Inscription 896'') and the question asks which country 
the person is in, but the answer (``Sweden'') appears 
nowhere in the context. Only LEB can transmit 
the sender's implicit landmark-to-country inference, 
yielding the largest contribution (+35.8\,pp). On 
multi-hop tasks (HotpotQA, 2WikiMQA, MuSiQue), 
LEB transmits reasoning chains integrated across 
documents (+19 to +22\,pp).

On MultiFieldQA, the question asks for a specific 
fact (e.g., ``What is the advantage of decorrelating 
the data before running the PLS algorithm?'') and the 
answer is a single extractable sentence located deep 
within a ${\sim}$7K-token document. No cross-entity 
reasoning is needed, and LEB adds only 
+7.0\,pp. This is also the sole task where KVComm 
\citep{shi2026kvcomm} outperforms \textsc{XBridge} 
(48.3 vs.\ 44.2): same-architecture KV sharing 
preserves position-level attention patterns that 
directly encode answer location, an advantage 
specific to retrieval tasks that disappears when 
reasoning is required (e.g., HotpotQA +13.5\,pp, 
MuSiQue +15.1\,pp over KVComm).

On Tipsheets, the answer is directly extractable from 
structured text, leaving minimal room for LEB 
improvement (+2.0\,pp).

\subsection{Multi-Agent Communication}
\label{appendix:multiagent}

\input{tables/tab_multiagent}

We test composability on HotpotQA, where each question requires reasoning over two supporting documents. We assign each document to a separate sender, Llama for the first and Mistral for the second, and let the receiver (Qwen) integrate both via their respective bridges. Each bridge was trained independently for its own model pair; no joint training or adaptation is performed.

Dual \textsc{XBridge} achieves 70.4\% F1 (Table~\ref{tab:multiagent}), outperforming both the single-sender full-context baseline (67.0\%, +3.4\,pp) and dual NLComm (56.8\%, +13.6\,pp). The single-sender baseline gives one Llama sender both documents, so the dual setup's improvement demonstrates that distributing documents across specialized senders is more effective than compressing everything through a single sender. This zero-shot composability, combining independently trained bridges without additional training, suggests that the framework scales naturally to multi-party agent settings.

\section{Additional Analysis}
\label{appendix:additional-analysis}

\subsection{Per-Sample Consistency}
\label{appendix:scatter}

The bridge mechanism analysis in 
Section~\ref{sec:mechanism-role} reports aggregate 
metrics across 196 HotpotQA samples 
(Llama$\to$Qwen). Here we verify that these 
improvements are consistent at the individual sample 
level rather than driven by a few outliers.

Figure~\ref{fig:mechanism-5cond}(c) plots the mean-centered cosine similarity between the generation representation and the answer entity representation for each sample, comparing NoComm and \textsc{XBridge}. 
Points above the $y = x$ line indicate samples where LEB improves alignment. 
Of 196 samples, 184 (94\%) fall above $y = x$ ($p \approx 0$, paired $t$-test). 
The population mean shifts from $\mu = -0.053$ (NoComm) to $\mu = +0.091$ (\textsc{XBridge}), a sign change indicating that without LEB the receiver's 
generation state moves away from the answer entity, while with LEB it moves toward it.

The 12 samples (6\%) where LEB does not improve 
alignment tend to involve short, unambiguous contexts 
where LAM alone already provides sufficient 
grounding. Even in these cases, LEB does not 
substantially degrade performance, indicating that the 
gating mechanism (Equation~\ref{eq:gated-residual}) 
successfully attenuates the bridge signal when it is 
not needed.

\subsection{Entity Perturbation Examples}
\label{appendix:perturbation-examples}

Table~\ref{tab:perturbation-examples} shows representative examples from the entity perturbation experiment (Section~\ref{sec:mechanism-role}). In each case, swapping LEB input alone does not change the output entity, while swapping LAM changes it entirely.

\input{tables/tab_perturbation_examples}

In all three examples, conditions A and D produce identical outputs, confirming that LEB does not influence entity selection. Condition C shifts the output to the swapped entity every time. LEB adapts its contextual enrichment accordingly: contextual signals such as ``port city 25\,km north of the museum'' or ``studied under in the Peripatetic school'' bind to whichever entity LAM provides.

\section{Task Details}
\label{appendix:tasks}

Table~\ref{tab:task-details} summarizes the seven benchmarks used in our evaluation. Training and evaluation splits are strictly separated with zero overlap.

\input{tables/tab_task_details}

\section{Related Work}
\label{sec:related-work}

\paragraph{Text-based multi-agent communication.}
LLMs are increasingly deployed in multi-agent systems where agents collaborate to solve tasks beyond individual capability \citep{guo2024multiagentsurvey, tran2025multi}. A parallel line asks what an agent should retain across interactions \citep{huang2026survey, huang2026harness}, whereas we focus on what is transferred within a single exchange. Multi-agent debate improves factuality by having models critique each other's reasoning \citep{du2024debate, liang2024encouraging}, while frameworks such as CAMEL \citep{li2023camel}, AutoGen \citep{wu2024autogen}, and ChatDev \citep{qian2024chatdev} coordinate agents through structured natural language exchanges. Natural language communication (NLComm) \citep{du2024debate} generalizes this pattern: one model generates a text summary and relays it to another. These text-based protocols are architecture-agnostic but require autoregressive decoding at every communication step, incurring high latency. They also discard the sender's internal representations, compressing context into a short summary that loses both the sender's contextual understanding and relational structure among entities.

\paragraph{Latent multi-agent communication.}
To avoid the cost and information loss of text-based communication, recent work
transmits internal model representations directly \citep{yu2026latentspace}. CIPHER \citep{pham2024cipher}
communicates via soft tokens across models but requires a shared tokenizer. AC
\citep{ramesh2025activation} shows that activation-space interventions transfer
between LLMs of the same family. C2C \citep{fu2025cache} projects and fuses KV
caches across model families through learned linear projections, achieving
cross-architecture transfer but operating entirely in continuous space. KVComm
\citep{shi2026kvcomm} shares selectively chosen KV cache layers, achieving
near-FullComm performance with reduced communication cost, but assumes matching
hidden dimensions, attention heads, and positional encodings.
These methods differ not only in what they transmit but in what they assume the
two models share. C2C fuses caches position-wise and therefore has the sharer and
receiver process the same context, and LatentMAS \citep{zou2025latentmas}
concatenates layer-wise KV caches across agents under a common architecture.
MoT \citep{feinashley2025mot} routes a query to a subset of frozen heterogeneous
experts and lets a designated primary expert cross-attend to its peers' projected
hidden states at uniformly placed interaction layers; all experts encode the same
query with their own tokenizers, so the sequences stay aligned and the primary
expert already holds the input. All three are symmetric in the sense that the
receiver already holds the context. XBridge
operates in the asymmetric setting of KVComm, where the sender observes a context
the receiver does not, and it is this asymmetry that gives rise to the entity
grounding problem: once the receiver can no longer see the entities directly, a
discrete channel aligned to its own vocabulary becomes necessary rather than
supplementary. Existing latent protocols therefore either assume architectural
homogeneity, require a shared tokenizer, or place the two models on the same
context, and none preserves discrete entity identity across different
vocabularies. XBridge addresses this gap by enabling latent communication across
different LLM architectures while preserving entity fidelity.

%% file: tables/tab_cross-vocab.tex
\begin{table}[ht]
\centering
\caption{Cross-vocabulary mapping statistics on HotpotQA
(100 validation samples). Direct-mapped tokens use ID-to-ID
lookup; the remainder use string-based fallback.}
\label{tab:vocab-stats}
\begin{tabular}{lcc}
\toprule
& Llama-3.1 $\to$ Qwen2.5 & Mistral-7B $\to$ Qwen2.5 \\
\midrule
Sender vocab size & 128{,}256 & 32{,}768 \\
Receiver vocab size & 151{,}665 & 151{,}665 \\
Shared tokens & 109{,}566 (85.4\%) & 10{,}566 (32.2\%) \\
\midrule
Direct-mapped & 97.1\% & 87.8\% \\
String fallback & 2.9\% & 12.2\% \\
Dominant mismatch type & Multi-digit numbers & Words + proper nouns \\
\bottomrule
\end{tabular}
\end{table}

%% file: tables/tab_vocab_example.tex
\begin{table}[ht]
\centering
\caption{Examples of cross-vocabulary mapping between different
tokenizer pairs. Direct lookup preserves the token as-is; string
fallback re-tokenizes the surface string, producing one-to-many
expansions.}
\label{tab:vocab-examples}
\begin{tabular}{llll}
\toprule
Source text & Sender tokens & Receiver tokens & Type \\
\midrule
\multicolumn{4}{l}{\textit{Llama-3.1 $\to$ Qwen2.5}} \\
\midrule
``the'' & [the] (1) & [the] (1) & Direct \\
``transformer'' & [transform, er] (2) & [transform, er] (2) & Direct \\
``199'' & [199] (1) & [1, 9, 9] (3) & Fallback \\
``COVID-19'' & [COVID, -, 19] (3) & [COVID, -, 1, 9] (4) & Mixed \\
\midrule
\multicolumn{4}{l}{\textit{Mistral-7B $\to$ Qwen2.5}} \\
\midrule
``directed'' & [directed] (1) & [direct, ed] (2) & Fallback \\
``Angeles'' & [Angeles] (1) & [An, ge, les] (3) & Fallback \\
``century'' & [century] (1) & [cent, ury] (2) & Fallback \\
\bottomrule
\end{tabular}
\end{table}

%% file: tables/tab_format_independence.tex
\begin{table}[t]
\centering
\caption{Grounding format comparison on HotpotQA, Llama$\to$Qwen (F1\,\%). 
Each grounding format is evaluated with and without LEB to isolate the bridge contribution ($\Delta$).}
\label{tab:format-independence}
\begin{tabular}{lccc}
\toprule
Grounding format & Without LEB & With LEB & LEB $\Delta$ \\
\midrule
LAM & 56.5 & 78.8 & +22.3 \\
NLComm text (chat prompt) & 68.5 & 81.0 & +12.5 \\
Argmax token map & 44.5 & 65.5 & +21.0 \\
\bottomrule
\end{tabular}
\end{table}

%% file: tables/tab_arch_ablation.tex
\begin{table}[t]
\centering
\caption{Bridge architecture ablations on HotpotQA, Llama$\to$Qwen (F1~\%). All variants use balanced training (587 samples).}
\label{tab:arch-ablation}
\small
\begin{tabular}{llc}
\toprule
Ablation & Variant & F1 \\
\midrule
\multirow{4}{*}{Modules} & 1 module (66M) & 66.3 \\
& 2 modules (132M) & 74.6 \\
& 4 modules (264M) & \textbf{78.8} \\
& 7 modules (462M) & 75.8 \\
\midrule
\multirow{4}{*}{Sender layer} & Layer 8 & 64.6 \\
& Layer 16 & 66.7 \\
& Layer 24 & 70.2 \\
& Layer 32 (last) & \textbf{78.8} \\
\midrule
\multirow{2}{*}{Gate init} & $\alpha = 0.0$ (Flamingo) & 72.7 \\
& $\alpha = 1.0$ (ours) & \textbf{78.8} \\
\midrule
\multirow{2}{*}{Loss} & NTP only (ours) & \textbf{78.8} \\
& NTP + contrastive & 78.7 \\
\bottomrule
\end{tabular}
\end{table}

%% file: tables/tab_data_composition.tex
\begin{table}[t]
\centering
\caption{Effect of training data composition on LEB performance (Llama$\to$Qwen, F1\,\%).}
\label{tab:data-composition}
\small
\begin{tabular}{lccccc}
\toprule
Bridge training & HotpotQA & Tipsheets & Countries & QASPER & Avg (7 tasks) \\
\midrule
HotpotQA only (20K) & 76.1 & 88.3 & 27.7 & 36.2 & 49.5 \\
Unbalanced (42K) & 77.1 & 100.0 & 66.0 & 45.3 & 59.3 \\
Balanced (587) & 78.8 & 99.8 & 72.5 & 47.5 & 63.2 \\
\bottomrule
\end{tabular}
\end{table}

%% file: tables/tab_homo_results.tex
\begin{table*}[t]
\centering
\caption{Same-architecture comparison (F1\,\%). \textsc{XBridge}$_\text{homo}$ uses Qwen2.5-7B as both sender and receiver with a separately trained bridge. FullComm = receiver reads full context directly (identical to Table~\ref{tab:main-results}). \textbf{Bold} = best per task (excluding FullComm). Improv.\ = improvement of \textsc{XBridge}$_\text{homo}$ over the second-best method (excluding FullComm).}
\label{tab:homo-results}
\small
\newcolumntype{C}[1]{>{\centering\arraybackslash}p{#1}}
\begin{tabular}{@{}l C{1.0cm} C{1.0cm} C{1.0cm} C{1.0cm} C{1.0cm} C{1.0cm} C{1.0cm} C{0.7cm}@{}}
\toprule
\textbf{Method} & \textbf{Countries} & \textbf{Tipsheets} & \textbf{HotpotQA} & \textbf{QASPER} & \textbf{MuSiQue} & \textbf{MFldQA} & \textbf{2Wiki} & \textbf{Avg} \\
\midrule
\multicolumn{9}{c}{\textit{$M_S$: Qwen2.5-7B-Instruct; $M_R$: Qwen2.5-7B-Instruct}} \\
\midrule
\textbf{FullComm} & 50.8 & 98.5 & 78.4 & 31.3 & 31.5 & 48.2 & 47.6 & 55.2 \\
\textbf{NLComm\textsubscript{homo}} & 37.2 & 96.0 & 64.9 & 30.8 & 29.1 & 23.9 & 18.2 & 42.9 \\
\textbf{KVComm} & 35.8 & 99.0 & 65.3 & 35.5 & 33.1 & \textbf{48.3} & 40.9 & 51.1 \\
\rowcolor{gray!15}
\textbf{\textsc{XBridge}$_\text{homo}$} & \textbf{72.7} & \textbf{99.8} & \textbf{87.1} & \textbf{49.0} & \textbf{38.7} & 45.6 & \textbf{48.9} & \textbf{63.1} \\%
\cdashline{1-9}\\[-8pt]
\quad Improv. & \small +35.5 & \small +0.8 & \small +21.8 & \small +13.5 & \small +5.6 & \small $-$2.7 & \small +8.0 & \small +12.0 \\
\bottomrule
\end{tabular}
\end{table*}

%% file: tables/bridge_contribution.tex
\begin{table}[t]
\centering
\caption{Bridge contribution per task 
(Llama$\to$Qwen, F1\,\%). 
\textsc{XBridge}\textsubscript{$\scalebox{0.8}{-\text{LEB}}$} = LAM 
only. $\Delta$ = \textsc{XBridge} $-$ 
\textsc{XBridge}\textsubscript{$\scalebox{0.8}{-\text{LEB}}$}.}
\label{tab:bridge-delta}
\small
\begin{tabular}{@{}llccc@{}}
\toprule
\textbf{Task} & \textbf{Type} 
& \textbf{\textsc{XBridge}\textsubscript{$\scalebox{0.8}{-\text{LEB}}$}} 
& \textbf{\textsc{XBridge}} 
& \textbf{$\Delta$} \\
\midrule
\textbf{Countries} & factual inference & 36.7 & 72.5 & +35.8 \\
\textbf{HotpotQA} & multi-hop QA & 56.5 & 78.8 & +22.3 \\
\textbf{2WikiMQA} & multi-hop QA & 29.5 & 51.1 & +21.6 \\
\textbf{MuSiQue} & compositional QA & 28.4 & 48.2 & +19.8 \\
\textbf{QASPER} & scientific QA & 33.0 & 47.5 & +14.5 \\
\textbf{MFldQA} & long-doc retrieval & 37.3 & 44.2 & +7.0 \\
\textbf{Tipsheets} & structured extract. & 97.8 & 99.8 & +2.0 \\
\bottomrule
\end{tabular}
\end{table}

%% file: tables/tab_multiagent.tex
\begin{table}[t]
\centering
\caption{Multi-agent communication on HotpotQA, Llama$\to$Qwen (F1~\%). Two senders each process one supporting document; the receiver integrates both.}
\label{tab:multiagent}
\begin{tabular}{lc}
\toprule
Configuration & F1 \\
\midrule
Single sender, full context (Llama) & 67.0 \\
Dual NLComm (Llama + Mistral) & 56.8 \\
Dual \textsc{XBridge} (Llama + Mistral) & \textbf{70.4} \\
\bottomrule
\end{tabular}
\end{table}

%% file: tables/tab_perturbation_examples.tex
\begin{table}[ht]
\centering
\caption{Representative entity perturbation examples on HotpotQA (Llama$\to$Qwen). The answer entity in the context is replaced with an unrelated entity of the same type. Condition A: both original. Condition D: LEB swapped only. Condition C: LAM swapped only.}
\label{tab:perturbation-examples}
\small
\resizebox{\linewidth}{!}{%
\begin{tabular}{@{}p{4.5cm}llccc@{}}
\toprule
\textbf{Question} & \textbf{Original} & \textbf{Swapped} & \textbf{A} & \textbf{D} & \textbf{C} \\
\midrule
Which port city lies $\sim$25\,km north of the Lingnan Fine Arts Museum? 
  & Keelung & Majuro & Keelung & Keelung & Majuro \\
\midrule
Who did Neleus of Scepsis study under in addition to Theophrastus?
  & Aristotle & Moroni & Aristotle & Aristotle & Moroni \\
\midrule
Who will Billy Howle be seen opposite in the upcoming British drama film directed by Dominic Cooke?
  & Saoirse Ronan & Kenji Mizoguchi & Saoirse Ronan & Saoirse Ronan & Kenji Mizoguchi \\
\bottomrule
\end{tabular}%
}
\end{table}

%% file: tables/tab_task_details.tex
\begin{table}[ht]
\centering
\caption{Task details. Context length statistics are measured across evaluation samples using the Llama-3.1 tokenizer.}
\label{tab:task-details}
\small
\begin{tabular}{@{}llcccc@{}}
\toprule
\textbf{Task} & \textbf{Type} & \textbf{Eval $N$} & \textbf{Median tok} & \textbf{Mean tok} & \textbf{Source} \\
\midrule
\textbf{HotpotQA} & multi-hop QA & 200 & 74 & 80 & \citet{yang2018hotpotqa} \\
\textbf{MuSiQue} & compositional QA & 200 & $\sim$1K$^*$ & $\sim$1.2K$^*$ & \citet{trivedi2022musique} \\
\textbf{QASPER} & scientific QA & 200 & 4,118 & 4,514 & \citet{dasigi2021qasper} \\
\textbf{2WikiMQA} & multi-hop QA & 100 & 6,532 & 7,267 & LongBench \citep{bai2024longbench} \\
\textbf{MultiFieldQA-en} & long-doc retrieval & 75 & 6,979 & 6,878 & LongBench \citep{bai2024longbench} \\
\textbf{Countries} & factual inference & 100 & 11 & 11 & \citet{shi2026kvcomm} \\
\textbf{Tipsheets} & structured extract. & 200 & 63 & 62 & \citet{shi2026kvcomm} \\
\bottomrule
\end{tabular}
\vspace{2pt}
\parbox{\linewidth}{\footnotesize $^*$MuSiQue statistics shown are for supporting paragraphs only; full context with distractors is approximately 4--5$\times$ longer. All evaluations use the full context.}
\end{table}

%% file: neurips_2026.bbl
\begin{thebibliography}{26}
\providecommand{\natexlab}[1]{#1}
\providecommand{\url}[1]{\texttt{#1}}
\expandafter\ifx\csname urlstyle\endcsname\relax
  \providecommand{\doi}[1]{doi: #1}\else
  \providecommand{\doi}{doi: \begingroup \urlstyle{rm}\Url}\fi

\bibitem[Zou et~al.(2025)Zou, Yang, Qiu, Li, Tieu, Lu, Shen, Tong, Choi, He,
  et~al.]{zou2025latentmas}
Jiaru Zou, Xiyuan Yang, Ruizhong Qiu, Gaotang Li, Katherine Tieu, Pan Lu,
  Ke~Shen, Hanghang Tong, Yejin Choi, Jingrui He, et~al.
\newblock Latent collaboration in multi-agent systems.
\newblock \emph{arXiv preprint arXiv:2511.20639}, 2025.

\bibitem[Hu et~al.(2025)Hu, Tan, Wang, Qu, and Chen]{hu2026multiagent}
Tianyu Hu, Zhen Tan, Song Wang, Huaizhi Qu, and Tianlong Chen.
\newblock Multi-agent debate for {LLM} judges with adaptive stability
  detection.
\newblock In \emph{The Thirty-ninth Annual Conference on Neural Information
  Processing Systems}, 2025.
\newblock URL \url{https://openreview.net/forum?id=Vusd1Hw2D9}.

\bibitem[Choi et~al.(2025)Choi, Zhu, and Li]{choi2026debate}
Hyeong~Kyu Choi, Jerry Zhu, and Sharon Li.
\newblock Debate or vote: Which yields better decisions in multi-agent large
  language models?
\newblock In \emph{The Thirty-ninth Annual Conference on Neural Information
  Processing Systems}, 2025.
\newblock URL \url{https://openreview.net/forum?id=iUjGNJzrF1}.

\bibitem[Du et~al.(2024)Du, Li, Torralba, Tenenbaum, and
  Mordatch]{du2024debate}
Yilun Du, Shuang Li, Antonio Torralba, Joshua~B Tenenbaum, and Igor Mordatch.
\newblock Improving factuality and reasoning in language models through
  multiagent debate.
\newblock In \emph{Forty-first international conference on machine learning},
  2024.

\bibitem[Shi et~al.(2026)Shi, Chiesa, Jr., and Kostic]{shi2026kvcomm}
Xiangyu Shi, Marco Chiesa, Gerald Q.~Maguire Jr., and Dejan Kostic.
\newblock {KVC}omm: Enabling efficient {LLM} communication through selective
  {KV} sharing.
\newblock In \emph{The Fourteenth International Conference on Learning
  Representations}, 2026.
\newblock URL \url{https://openreview.net/forum?id=F7rUng23nw}.

\bibitem[Ramesh and Li(2025)]{ramesh2025activation}
Vignav Ramesh and Kenneth Li.
\newblock Communicating activations between language model agents.
\newblock In \emph{Forty-second International Conference on Machine Learning},
  2025.
\newblock URL \url{https://openreview.net/forum?id=W6RPXUUFic}.

\bibitem[Fu et~al.(2026)Fu, Min, Zhang, Yan, Dai, Ouyang, and
  Wang]{fu2025cache}
Tianyu Fu, Zihan Min, Hanling Zhang, Jichao Yan, Guohao Dai, Wanli Ouyang, and
  Yu~Wang.
\newblock Cache-to-cache: Direct semantic communication between large language
  models.
\newblock In \emph{The Fourteenth International Conference on Learning
  Representations}, 2026.
\newblock URL \url{https://openreview.net/forum?id=LeatkxrBCi}.

\bibitem[Pham et~al.(2024)Pham, Liu, Yang, Chen, Liu, Yuan, Plummer, Wang, and
  Yang]{pham2024cipher}
Chau Pham, Boyi Liu, Yingxiang Yang, Zhengyu Chen, Tianyi Liu, Jianbo Yuan,
  Bryan~A. Plummer, Zhaoran Wang, and Hongxia Yang.
\newblock Let models speak ciphers: Multiagent debate through embeddings.
\newblock In \emph{The Twelfth International Conference on Learning
  Representations}, 2024.
\newblock URL \url{https://openreview.net/forum?id=sehRvaIPQQ}.

\bibitem[Alayrac et~al.(2022)Alayrac, Donahue, Luc, Miech, Barr, Hasson, Lenc,
  Mensch, Millican, Reynolds, et~al.]{alayrac2022flamingo}
Jean-Baptiste Alayrac, Jeff Donahue, Pauline Luc, Antoine Miech, Iain Barr,
  Yana Hasson, Karel Lenc, Arthur Mensch, Katherine Millican, Malcolm Reynolds,
  et~al.
\newblock Flamingo: a visual language model for few-shot learning.
\newblock \emph{Advances in neural information processing systems},
  35:\penalty0 23716--23736, 2022.

\bibitem[Lee et~al.(2026)Lee, Yang, and Kang]{lee2024subgraph}
Junhyun Lee, Wooseong Yang, and Jaewoo Kang.
\newblock Subgraph-level universal prompt tuning.
\newblock \emph{Information Sciences}, page 123516, 2026.

\bibitem[Grattafiori et~al.(2024)Grattafiori, Dubey, Jauhri, Pandey, Kadian,
  Al-Dahle, Letman, Mathur, Schelten, Vaughan, et~al.]{dubey2024llama3}
Aaron Grattafiori, Abhimanyu Dubey, Abhinav Jauhri, Abhinav Pandey, Abhishek
  Kadian, Ahmad Al-Dahle, Aiesha Letman, Akhil Mathur, Alan Schelten, Alex
  Vaughan, et~al.
\newblock The llama 3 herd of models.
\newblock \emph{arXiv preprint arXiv:2407.21783}, 2024.

\bibitem[{Qwen Team}(2024)]{qwen2024qwen25}
{Qwen Team}.
\newblock Qwen2.5 technical report.
\newblock \emph{arXiv preprint arXiv:2412.15115}, 2024.

\bibitem[Yang et~al.(2018)Yang, Qi, Zhang, Bengio, Cohen, Salakhutdinov, and
  Manning]{yang2018hotpotqa}
Zhilin Yang, Peng Qi, Saizheng Zhang, Yoshua Bengio, William Cohen, Ruslan
  Salakhutdinov, and Christopher~D Manning.
\newblock Hotpotqa: A dataset for diverse, explainable multi-hop question
  answering.
\newblock In \emph{Proceedings of the 2018 conference on empirical methods in
  natural language processing}, pages 2369--2380, 2018.

\bibitem[Trivedi et~al.(2022)Trivedi, Balasubramanian, Khot, and
  Sabharwal]{trivedi2022musique}
Harsh Trivedi, Niranjan Balasubramanian, Tushar Khot, and Ashish Sabharwal.
\newblock Musique: Multihop questions via single-hop question composition.
\newblock \emph{Transactions of the Association for Computational Linguistics},
  10:\penalty0 539--554, 2022.

\bibitem[Dasigi et~al.(2021)Dasigi, Lo, Beltagy, Cohan, Smith, and
  Gardner]{dasigi2021qasper}
Pradeep Dasigi, Kyle Lo, Iz~Beltagy, Arman Cohan, Noah~A Smith, and Matt
  Gardner.
\newblock A dataset of information-seeking questions and answers anchored in
  research papers.
\newblock In \emph{Proceedings of the 2021 Conference of the North American
  Chapter of the Association for Computational Linguistics: Human Language
  Technologies}, pages 4599--4610, 2021.

\bibitem[Bai et~al.(2024)Bai, Lv, Zhang, Lyu, Tang, Huang, Du, Liu, Zeng, Hou,
  et~al.]{bai2024longbench}
Yushi Bai, Xin Lv, Jiajie Zhang, Hongchang Lyu, Jiankai Tang, Zhidian Huang,
  Zhengxiao Du, Xiao Liu, Aohan Zeng, Lei Hou, et~al.
\newblock Longbench: A bilingual, multitask benchmark for long context
  understanding.
\newblock In \emph{Proceedings of the 62nd annual meeting of the association
  for computational linguistics (volume 1: Long papers)}, pages 3119--3137,
  2024.

\bibitem[Guo et~al.(2024)Guo, Chen, Wang, Chang, Pei, Chawla, Wiest, and
  Zhang]{guo2024multiagentsurvey}
Taicheng Guo, Xiuying Chen, Yaqi Wang, Ruidi Chang, Shichao Pei, Nitesh~V.
  Chawla, Olaf Wiest, and Xiangliang Zhang.
\newblock Large language model based multi-agents: a survey of progress and
  challenges.
\newblock In \emph{Proceedings of the Thirty-Third International Joint
  Conference on Artificial Intelligence}, IJCAI '24, 2024.
\newblock ISBN 978-1-956792-04-1.
\newblock \doi{10.24963/ijcai.2024/890}.
\newblock URL \url{https://doi.org/10.24963/ijcai.2024/890}.

\bibitem[Tran et~al.(2025)Tran, Dao, Nguyen, Pham, O'Sullivan, and
  Nguyen]{tran2025multi}
Khanh-Tung Tran, Dung Dao, Minh-Duong Nguyen, Quoc-Viet Pham, Barry O'Sullivan,
  and Hoang~D Nguyen.
\newblock Multi-agent collaboration mechanisms: A survey of llms.
\newblock \emph{arXiv preprint arXiv:2501.06322}, 2025.

\bibitem[Huang et~al.(2026{\natexlab{a}})Huang, Zhang, Liang, Bei, Chen, Feng,
  Tan, Wang, Wei, Wu, et~al.]{huang2026survey}
Wei-Chieh Huang, Weizhi Zhang, Yueqing Liang, Yuanchen Bei, Yankai Chen, Tao
  Feng, Zhen Tan, Yu~Wang, Tianxin Wei, Shanglin Wu, et~al.
\newblock A survey of agent memory in the second half: Towards self-evolving
  and long-horizon agents.
\newblock \emph{Transactions on Machine Learning Research}, 2026{\natexlab{a}}.

\bibitem[Huang et~al.(2026{\natexlab{b}})Huang, Zhang, Wu, Chen, Jiang, Yang,
  Yang, Zou, Zhang, Wu, et~al.]{huang2026harness}
Wei-Chieh Huang, Weizhi Zhang, Yuchen Wu, Yankai Chen, Eric~Hanchen Jiang,
  Wooseong Yang, Yiwei Yang, Henry~Peng Zou, Hanrong Zhang, Ying~Nian Wu,
  et~al.
\newblock Harness the memory: A holistic evaluation of memory substrates in
  memory agents.
\newblock \emph{arXiv preprint arXiv:2608.15008}, 2026{\natexlab{b}}.

\bibitem[Liang et~al.(2024)Liang, He, Jiao, Wang, Wang, Wang, Yang, Shi, and
  Tu]{liang2024encouraging}
Tian Liang, Zhiwei He, Wenxiang Jiao, Xing Wang, Yan Wang, Rui Wang, Yujiu
  Yang, Shuming Shi, and Zhaopeng Tu.
\newblock Encouraging divergent thinking in large language models through
  multi-agent debate.
\newblock In \emph{Proceedings of the 2024 conference on empirical methods in
  natural language processing}, pages 17889--17904, 2024.

\bibitem[Li et~al.(2023)Li, Hammoud, Itani, Khizbullin, and
  Ghanem]{li2023camel}
Guohao Li, Hasan Hammoud, Hani Itani, Dmitrii Khizbullin, and Bernard Ghanem.
\newblock Camel: Communicative agents for" mind" exploration of large language
  model society.
\newblock \emph{Advances in neural information processing systems},
  36:\penalty0 51991--52008, 2023.

\bibitem[Wu et~al.(2024)Wu, Bansal, Zhang, Wu, Li, Zhu, Jiang, Zhang, Zhang,
  Liu, et~al.]{wu2024autogen}
Qingyun Wu, Gagan Bansal, Jieyu Zhang, Yiran Wu, Beibin Li, Erkang Zhu,
  Li~Jiang, Xiaoyun Zhang, Shaokun Zhang, Jiale Liu, et~al.
\newblock Autogen: Enabling next-gen llm applications via multi-agent
  conversations.
\newblock In \emph{First conference on language modeling}, 2024.

\bibitem[Qian et~al.(2024)Qian, Liu, Liu, Chen, Dang, Li, Yang, Chen, Su, Cong,
  et~al.]{qian2024chatdev}
Chen Qian, Wei Liu, Hongzhang Liu, Nuo Chen, Yufan Dang, Jiahao Li, Cheng Yang,
  Weize Chen, Yusheng Su, Xin Cong, et~al.
\newblock Chatdev: Communicative agents for software development.
\newblock In \emph{Proceedings of the 62nd annual meeting of the association
  for computational linguistics (volume 1: Long papers)}, pages 15174--15186,
  2024.

\bibitem[Yu et~al.(2026)Yu, Chen, He, Fu, Dong, Yang, Xu, Ma, Hu, Cao,
  et~al.]{yu2026latentspace}
Xinlei Yu, Zhangquan Chen, Yongbo He, Tianyu Fu, Guanting Dong, Cheng Yang,
  Chengming Xu, Yue Ma, Xiaobin Hu, Zhe Cao, et~al.
\newblock The latent space: Foundation, evolution, mechanism, ability, and
  outlook.
\newblock \emph{arXiv preprint arXiv:2604.02029}, 2026.

\bibitem[Fein-Ashley et~al.(2025)Fein-Ashley, Parikh, Kannan, and
  Prasanna]{feinashley2025mot}
Jacob Fein-Ashley, Dhruv Parikh, Rajgopal Kannan, and Viktor Prasanna.
\newblock Mixture of thoughts: Learning to aggregate what experts think, not
  just what they say.
\newblock \emph{arXiv preprint arXiv:2509.21164}, 2025.

\end{thebibliography}
